\documentclass[11pt]{article}
\IfFileExists{koa_2/main.tex}
  {\providecommand{\koaroot}{koa_2/}}
  {\providecommand{\koaroot}{}}
\makeatletter
\edef\input@path{{\koaroot}}
\makeatother

\usepackage[letterpaper,margin=1in,headheight=14pt]{geometry}

\usepackage[utf8]{inputenc}
\usepackage[T1]{fontenc}
\usepackage{amsmath}
\usepackage{amssymb}
\usepackage{amsfonts}
\usepackage[scaled=0.92]{helvet}   % Helvetica-like sans for headings
\usepackage{microtype}
\usepackage{nicefrac}

\usepackage{graphicx}
\usepackage{xcolor}
\usepackage{colortbl}
\usepackage{booktabs}
\usepackage{tabularx}
\usepackage{array}
\usepackage{adjustbox}
\usepackage{rotating}
\usepackage{float}
\usepackage{placeins}
\usepackage{longtable}
\usepackage[font=small,justification=raggedright,singlelinecheck=false]{caption}
\usepackage{enumitem}
\usepackage{listings}
\usepackage{tikz}
\usetikzlibrary{arrows.meta,calc}
\newcolumntype{Y}{>{\raggedright\arraybackslash}X}

\usepackage[numbers,compress]{natbib}

\definecolor{koablue}{RGB}{0,112,210}     % Salesforce accent blue
\definecolor{koadark}{RGB}{21,55,102}     % deep navy for headings
\definecolor{koagray}{RGB}{90,90,90}
\definecolor{koawin}{RGB}{209,231,221}    % soft green highlight for our rows
\definecolor{koareg}{RGB}{250,235,215}    % soft amber highlight for regressions

\usepackage{hyperref}
\usepackage{url}
\hypersetup{
  colorlinks=true,
  linkcolor=koadark,
  citecolor=koablue,
  urlcolor=koablue,
  pdftitle={Salesforce Koa: An Enterprise Language Model for Agentic Tool Use},
  pdfauthor={Salesforce Agentforce and AI Research (Koa Team)}
}

\makeatletter
\renewcommand\section{\@startsection{section}{1}{\z@}%
  {-1.6ex \@plus -.4ex \@minus -.2ex}%
  {0.9ex \@plus .2ex}%
  {\normalfont\Large\bfseries\sffamily\color{koadark}}}
\renewcommand\subsection{\@startsection{subsection}{2}{\z@}%
  {-1.3ex \@plus -.3ex \@minus -.2ex}%
  {0.7ex \@plus .2ex}%
  {\normalfont\large\bfseries\sffamily\color{koadark}}}
\renewcommand\subsubsection{\@startsection{subsubsection}{3}{\z@}%
  {-1.1ex \@plus -.3ex \@minus -.2ex}%
  {0.5ex \@plus .2ex}%
  {\normalfont\normalsize\bfseries\sffamily\color{koadark}}}
\makeatother

\usepackage{fancyhdr}
\fancypagestyle{plain}{%
  \fancyhf{}\fancyfoot[C]{\thepage}}

\newcommand{\code}[1]{\texttt{#1}}
\newcommand{\ours}{\rowcolor{koawin}}
\newcommand\blfootnote[1]{%
  \begingroup\renewcommand\thefootnote{}\footnote{#1}%
  \addtocounter{footnote}{-1}\endgroup}

\begin{document}

% ---- Title block ----------------------------------------------------------
\thispagestyle{plain}

\begin{center}
  {\LARGE\bfseries\color{koadark}
   Salesforce Koa:\\[0.35em]
   An Enterprise Language Model for Agentic Tool Use\par}
  \vspace{1.3em}
  \begin{minipage}{\textwidth}
  \centering
  {\normalsize
   Zixiang~Chen\textsuperscript{*}, Sufeng~Niu\textsuperscript{*},
   Yingchi~Liu\textsuperscript{*}, Wenting~Zhao\textsuperscript{*},
   Akshara~Prabhakar\textsuperscript{*}, Shubham~Mehrotra\textsuperscript{*}\par}
  \end{minipage}\par
  \vspace{0.7em}
  \begin{minipage}{0.92\textwidth}
  \centering
  {\normalsize
   Bin~Bi, \enspace Zhujun~Lan, \enspace Katherine~Tan, \enspace
   Mohammad~Ramezanali, \enspace Tulika~Manoj~Awalgaonkar, \enspace
   Monojit~Banerjee, \enspace Jielin~Qiu, \enspace Shiva~Kumar~Pentyala, \enspace
   Zhepeng~Cen, \enspace Anupam~Tripathi, \enspace Ali~Ziaei, \enspace
   Regunathan~Radhakrishnan\par}
  \vspace{0.7em}
  {\normalsize
   Darvish~Lee~Shadravan, \enspace Shelby~Heinecke, \enspace Sitaram~Asur, \enspace
   Jayesh~Govindarajan, \enspace Silvio~Savarese, \enspace James~Zhu, \enspace
   Phil~Mui\textsuperscript{\dag}, \enspace Huan~Wang\textsuperscript{\dag}\par}
  \end{minipage}
  \blfootnote{\textsuperscript{*}\,Equal contribution (co-first authors).
    \hspace{1em}\textsuperscript{\dag}\,Co-corresponding authors.}
  \vspace{1.0em}\par
  {\large Salesforce Agentforce \& AI Research\par}
  \vspace{0.9em}
  {\color{koagray}\normalsize\today\par}
\end{center}
\vspace{0.6em}

% ---- Front matter & body sections ----------------------------------------
\begin{abstract}
We present Salesforce Koa, an enterprise language model built by post-training the
open-weight Nemotron-3-Super-120B foundation model with reinforcement learning
using Group Relative Policy Optimization (GRPO), and deployed in FP8 for
production. Koa is trained only on public and synthetically generated data, and
specialized for the agentic tool use that enterprise workflows
demand: routing a request to the correct action, invoking the right tool with valid
arguments, and completing multi-turn business tasks. The distinctive component of
our pipeline is specification-driven task construction: declarative Agent Script
specifications are expanded into persona-conditioned multi-turn environments whose
rewards are grounded in successful tool use. Applied to enterprise CRM
specifications, the same pipeline produces the in-domain training distribution on
which Koa is specialized. On CRMAgentBench and the human-labeled production
tool-calling set, Koa outperforms both its untuned
open-weight base and GPT-4.1 and is competitive with the strongest frontier models.
It reaches 87\% Task Success Rate on
CRMAgentBench (vs.\ GPT-4.1 at 82\% and the base at 79\%), is at or near the top of
every metric on the human-labeled portion of an internal production benchmark,
and preserves the base model's general capability on public
benchmarks (Tau2Bench, BFCL). A controlled comparison with architecture and RL
recipe held fixed shows that the additional in-domain RL stage improves argument
accuracy and full tool-call success on the human-labeled enterprise benchmark.
\end{abstract}

\section{Introduction}
\label{sec:intro}
Large language models (LLMs) increasingly power enterprise AI systems, where,
unlike general-purpose assistants, they must invoke external tools, retrieve
structured information, and complete multi-step workflows across business
applications. Robust tool use and agentic reasoning are therefore the essential
enterprise capabilities. Many production systems rely on proprietary frontier
models, yet enterprises often need more control over deployment, customization,
governance, and cost than a single external provider allows. Recent open-weight
foundation models such as Llama~\cite{dubey2024llama3},
Gemma~\cite{gemmateam2025gemma3}, Qwen~\cite{yang2024qwen2},
Mistral~\cite{jiang2023mistral}, and Nemotron~\cite{nvidia2024nemotron} have
closed much of the public-benchmark gap, and access to weights lets organizations
post-train for their own domains. A central question is whether they can be
effectively specialized for enterprise agentic tasks while preserving
general-purpose capability, and, if so, \emph{what} matters most in doing so.

We introduce Salesforce Koa, an enterprise language model built by post-training
the open-weight Nemotron-3-Super-120B foundation model (Nemotron-120B) with GRPO,
using only public and synthetically generated data, and served
in FP8 for production. The distinctive component of our pipeline is
specification-driven task construction: for enterprise domains we build training
tasks from declarative agent specifications written in Agent
Script~\cite{salesforce_agentscript}, Salesforce's declarative language for
Agentforce agents. A specification captures an agent's routing structure,
subagents, typed actions, tool scopes, and workflow instructions; a simulation
pipeline expands it into scenario- and persona-conditioned tasks that NeMo
Gym~\cite{nemo-gym} executes as online rollouts, with grounded task-resolution
rewards driving GRPO.
This creates a direct bridge between agent authoring and model post-training: the
same specifications that configure an agent also structure its rollout tasks and
resolution criteria.

We organize the paper around the progression that produced Koa: from the untuned base, to RL on public and synthetic multi-domain data, to RL on the in-domain Salesforce distribution, to the FP8 model we deploy. RL on public and synthetic data yields modest gains on public benchmarks. Continuing training on the in-domain Salesforce distribution, with architecture and RL recipe held fixed, further improves argument accuracy and full tool-call success on the human-labeled enterprise benchmark. FP8 quantization yields comparable aggregate benchmark performance while making the 120B model efficient to serve.

Because Koa is specialized for a specific production distribution, we evaluate it
primarily on that distribution: we lead with enterprise benchmarks that reflect
enterprise workflows---CRMAgentBench and a production tool-calling
benchmark---and use public tool-calling benchmarks such as
BFCL~\cite{patil2025bfcl} as a check that general-purpose capability is preserved.
Reporting both lets us separate what enterprise specialization changes on the
deployment distribution from what it changes on generalist tasks.

Our contributions are as follows. First, we present Koa, an enterprise model
post-trained from the open-weight Nemotron-120B base with a specification-driven,
simulation-to-reward RL pipeline and deployed in FP8; on CRMAgentBench it reaches
87\% Task Success Rate, surpassing GPT-4.1 (82\%) and the untuned Nemotron base
(79\%) and matching Nemotron-Ultra-550B, and on the human-labeled
set of an internal production tool-calling benchmark it achieves the highest
non-hallucination rate of any model evaluated and higher tool-name accuracy
than all evaluated open-weight baselines. 
Second, holding architecture and RL recipe fixed, we compare the public-data RL checkpoint with the checkpoint obtained through additional in-domain RL, showing improved argument accuracy and full tool-call success on the human-labeled enterprise benchmark.
Third, we
provide a capability-preservation analysis on public benchmarks, including a
per-subset account of a small BFCL change. Finally, we report a deployment study showing that
FP8 quantization with in-domain calibration retains comparable aggregate benchmark performance for
efficient 120B serving.

\section{Related Work}
\label{sec:related}

\paragraph{Open-weight foundation models.}
Recent advances in open-weight foundation models have substantially narrowed the
gap with proprietary frontier models across reasoning, coding, and agentic tasks.
Model families such as Llama~\cite{dubey2024llama3}, Gemma~\cite{gemmateam2025gemma3},
Qwen~\cite{yang2024qwen2}, Mistral~\cite{jiang2023mistral}, and
Nemotron~\cite{nvidia2024nemotron} show that open weights provide a strong
foundation for downstream adaptation, letting organizations post-train on their
own data while retaining control over deployment and customization. We build on an
open-weight Nemotron foundation model and study reinforcement learning as a
mechanism for enterprise specialization.

\paragraph{Enterprise language models and tool use.}
LLMs are increasingly deployed as agents that interact with external tools rather
than only generating text. Toolformer~\cite{schick2023toolformer},
Gorilla~\cite{patil2024gorilla}, and ToolLLM~\cite{qin2024toolllm} demonstrate API
invocation, function calling, and tool use, and benchmarks such as
BFCL~\cite{patil2025bfcl}, CRMArena~\cite{huang-etal-2025-crmarena}, and
LoCoBench-Agent~\cite{qiu2025locobenchagent} establish tool use and agentic
reasoning as critical capabilities. A complementary line of work builds
\emph{large action models} together with the data and tooling to train them:
xLAM~\cite{zhang2025xlam} trains open action-model families for agentic tasks;
APIGen~\cite{liu2024apigenautomatedpipelinegenerating} and
APIGen-MT~\cite{prabhakar2025apigenmt} synthesize verifiable function-calling and
multi-turn agent data; ActionStudio~\cite{zhang2025actionstudio} provides a
lightweight training framework for such models; and
ToolLibGen~\cite{yue2025toollibgen} automatically creates and aggregates tool
libraries. Enterprise applications
place unique demands on these capabilities: rather than answering standalone
questions, enterprise agents must retrieve structured information, invoke business
APIs, and execute multi-step workflows over organization-specific schemas and
policies, where tools may have overlapping functionality and an incorrect action
can affect live business records. We focus on this enterprise setting and study
how RL specializes an open-weight model for it, evaluating on both public
function-calling benchmarks and enterprise CRM tasks.

\paragraph{Post-training for LLMs.}
Post-training is a critical stage in building capable models. Supervised
fine-tuning (SFT) adapts foundation models to downstream tasks and improves
instruction following~\cite{ouyang2022instructgpt}, while reinforcement
learning---from RLHF and Constitutional AI~\cite{bai2022constitutional} to
policy-gradient methods such as PPO~\cite{schulman2017ppo} and Group Relative
Policy Optimization (GRPO)~\cite{shao2024deepseekmath,guo2025deepseekr1},
preference optimization~\cite{rafailov2023dpo}, and on-policy
distillation~\cite{agarwal2024onpolicy}---further improves reasoning and
decision-making. A growing body of work continues to scale and specialize RL
post-training~\cite{cen2026webscalerl,qian2025userrl}.

Turn-local RL methods focus updates on selected agent decisions.
PivotRL~\cite{yi2026pivotrl} filters intermediate turns from expert trajectories
using verifier reward statistics and performs local RL updates.
Critical-State RL~\cite{chen2026criticalstate} checks reward alignment and
improvement headroom, separates action-dependent reward variation from
continuation noise through nested sampling, and trains the selected model calls.
We study post-training for enterprise adaptation from an \emph{already}
post-trained open-weight foundation model, comparing SFT and RL and, more
centrally, isolating the effect of training on the deployment distribution. Our results show that additional in-domain RL improves argument accuracy and full tool-call success on the human-labeled enterprise benchmark.

\section{Method}
\label{sec:method}

Salesforce Koa is built by post-training the open-weight Nemotron-120B foundation
model with GRPO. Training uses only public resources and synthetically generated
interactions. The distinctive component of our pipeline
is \emph{specification-driven task construction}: declarative enterprise agent
specifications are expanded into executable, persona-conditioned multi-turn
environments whose rewards are grounded in successful tool use; instantiated on
production Salesforce specifications, the same construction yields the in-domain
distribution on which Koa is specialized. We describe environment and task construction
(Section~\ref{sec:env-authoring}), online rollout and the grounded reward
(Section~\ref{sec:environment}), policy optimization (Section~\ref{sec:rl-opt}),
the training-data regimes that define our central comparison
(Section~\ref{sec:data-regimes}), and FP8 quantization for deployment
(Section~\ref{sec:fp8}). We apply RL directly to the base model rather than to an
SFT checkpoint; a preliminary SFT study that motivated this choice, together with
distributed-training and evaluation details, is deferred to
Appendix~\ref{appx:sft} and~\ref{appx:infra}.

\subsection{Spec-driven environment and task construction}
\label{sec:env-authoring}

For enterprise domains, we author workflow structure in Agent Script. A
specification defines a router, specialized subagents, typed actions, and
natural-language reasoning instructions, which together determine how requests
are routed, which tools are exposed within each topic, and how each topic handles
its workflow. A static extraction pass compiles this specification into a typed
workflow graph capturing per-tool argument schemas, declared state effects,
routing conditions, and loop/termination configuration; this graph instantiates
the executable environment, including the simulated state that tools read and
mutate. For public tool-use domains, where no Agent Script specification exists,
we instead synthesize the workflow structure directly, producing the same typed
workflow graph of tools, argument schemas, and termination conditions; both
sources therefore feed the shared simulation and reward machinery below equally.

A simulation pipeline expands each specification into scenario- and
persona-conditioned sessions: the scenario determines the workflows and user
needs exercised, while the persona conditions the simulated customer's behavior.
These sessions are converted into RL examples that begin either from the opening
request or from a generated dialogue prefix, exposing the policy to decisions at
different depths of a multi-turn interaction. Each example uses a common schema
in which \code{agent\_ref} selects the environment and the remaining fields
provide the initial rollout context (task intent, persona, dialogue state,
available tools); the full trajectory is generated online. When multiple
environments are trained jointly, per-row \code{agent\_ref} dispatch lets a
single run load them all, and differential exposure is realized by repeating an
environment's examples before shuffling. NeMo Gym then executes each example as
an online rollout, so workflow authoring and task construction are cleanly
separated from rollout execution and policy optimization.

\subsection{Online rollout and grounded reward}
\label{sec:environment}

Each example is executed inside a NeMo Gym environment. For simulated-dialogue
environments a single frozen helper model plays three roles in every rollout: a
customer simulator (produces the next user turn from the sampled persona),
a tool/function emulator (produces tool outputs when no production backend
exists), and a coverage judge (scores resolution to form the reward). The
helper is a separate inference service and is never updated by policy
optimization. Figure~\ref{fig:rollout} in Appendix~\ref{appx:rollout} traces one
rollout end to end, making explicit which components are frozen versus trained,
the order of interactions within a turn, and where the scalar reward comes from.

The reward is built from the coverage judge. The task intent is represented as
numbered sub-questions, and the judge emits one resolved/not-resolved verdict per
sub-question over the full transcript. The coverage rate is
\begin{equation}
  \mathrm{cov}
  =
  \frac{\#\{\text{sub-questions judged resolved}\}}{\#\{\text{sub-questions}\}}
  \in [0,1],
  \label{eq:coverage}
\end{equation}
where missing or malformed verdicts default to not resolved. The scalar reward
applies a duplicate-call gate,
\begin{equation}
  R
  =
  \begin{cases}
    0, & \text{a consecutive duplicate tool call occurred,}\\
    \mathrm{cov}, & \text{otherwise,}
  \end{cases}
  \label{eq:reward}
\end{equation}
and the judge prompt requires that any sub-question needing customer- or
system-specific data be marked resolved only when the answer is grounded in a
successful relevant tool call; generic factoids may resolve without a tool. The
deterministic sandbox environment instead sets $R=1$ exactly when the predicted
actions reproduce the reference end state and $R=0$ otherwise. This grounding is
deliberate: it rewards the model for \emph{acting} through the right tool, the
behavior enterprise workflows require, and---as we discuss in
Section~\ref{sec:analysis}---also shapes which public-benchmark categories move
after specialization.

A few properties of these environments are central to their behavior. The tool
list is per rollout, not global: each conversation advertises only the
tools available in its source session, and in routed environments the visible set
is further restricted to the active subagent, so the policy sees only
in-scope tools at any moment. Task intent is constructed deterministically at
build time (a numbered list of sub-questions), directly connecting task
construction to the reward. A persona is sampled per conversation and held
constant, controlling how the simulated user reacts, escalates, or declares
resolution. Focus system prompts are not restricted to the conversation
start: in routed environments each subagent switch appends a fresh
focus/procedure message, re-focusing the policy exactly when the topic changes.
Environments differ mainly in interaction topology and verification
(Appendix~\ref{appx:env-summary}, Table~\ref{tab:env-summary}).

\subsection{Optimization}
\label{sec:rl-opt}

We maximize the expected trajectory reward
$\mathcal{J}(\theta)=\mathbb{E}_{x\sim\mathcal{D}}\,
\mathbb{E}_{\tau\sim\pi_\theta(\cdot\mid x,\mathrm{env})}[R(\tau)]$
with GRPO variants: For each prompt the policy samples a small group of complete
trajectories with colocated inference; dynamic sampling discards zero-variance
groups and refills until a full batch is assembled. Advantages are estimated by a
leave-one-out group baseline (no value network), and a token-level truncated
importance-sampling weight corrects the train/generation log-probability
mismatch under a single on-policy update per batch. Trajectories with malformed
tool-call or thinking syntax have the offending token advantages set to a fixed
negative value, and over-length or log-probability-inconsistent trajectories are
masked. The full objective, advantage normalization, and validity constraints
are given in Appendix~\ref{appx:grpo}. A complementary single-step training mode
for constrained decision turns, which isolates the rare but decisive turns where
the agent must communicate with the user rather than call a tool, is described in
Appendix~\ref{appx:constrained}.

\subsection{Training-data regimes}
\label{sec:data-regimes}

Koa is produced by two RL stages on data from different regimes, applied in
sequence from the Nemotron base, with the RL recipe of
Sections~\ref{sec:environment}--\ref{sec:rl-opt} held fixed throughout. The first
stage uses the \textbf{public / synthetic} regime, which generates rollouts against
multi-domain environments (\code{salesforce support}, \code{healthcare
administration}, \code{real estate}, and a helper-free \code{workplace assistant}
sandbox) built from public tool schemas and synthesized workflow structure,
measuring how far a general spec-driven recipe carries on \emph{public} benchmarks.
We then continue training with a second stage on the \textbf{in-domain Salesforce}
regime, which instantiates the same spec-driven construction
(Section~\ref{sec:env-authoring}) on production Salesforce agent specifications and
tool schemas, so that the synthesized sessions' tool inventories, argument
conventions, and multi-turn workflows match the production distribution the model
will serve; as in the public/synthetic regime, it uses only specification-derived
and synthesized interactions. This in-domain stage produces the deployed Koa
checkpoint. Because the base and RL recipe are held fixed and the only change is the
added in-domain stage, the improvement on the production distribution is
attributable to in-domain adaptation. Section~\ref{sec:internal-benchmarks} reports
the controlled result; Section~\ref{sec:analysis} discusses the effects of in-domain adaptation and FP8 deployment.

\subsection{FP8 quantization for deployment}
\label{sec:fp8}

The deployed Koa checkpoint is served in FP8 for inference efficiency. We apply
post-training quantization (PTQ) with NVIDIA ModelOpt, using \emph{in-domain}
calibration data drawn from the production distribution rather than generic
calibration text, so that quantization statistics reflect the tool-calling
traffic the model actually serves. We refer to the unquantized checkpoint as
\textbf{Koa (BF16)} and the deployed model as \textbf{Koa (FP8)}, and
report both so the effect of quantization can be isolated
(Section~\ref{sec:analysis}).

\section{Experiments}
\label{sec:experiments}

We post-train the Nemotron-3-Super-120B policy on multi-domain agentic
tool-calling tasks and evaluate it against strong open and proprietary baselines.
Consistent with the motivation in Section~\ref{sec:related}, we \emph{lead with
enterprise benchmarks} (Section~\ref{sec:internal-benchmarks}), which reflect the
production distribution Koa is built to serve, and then report public benchmarks
(Section~\ref{sec:public-benchmarks}) as a check that general capability is
preserved.

\subsection{Setup}
\label{sec:setup}

\paragraph{Policies and helper.}
Both trained policies are Nemotron reasoning models run in thinking mode. The
Nemotron-3-Super-120B is the final target policy; a Nano~v3 ($\sim$30B) model
served as a cheap proxy on which we developed the GRPO reward/algorithm recipe
before porting it to the 120B policy, and the same Nano-30B runs inside the
environments as the frozen helper that plays customer simulator, tool
emulator, and coverage judge.

\begin{table}[t]
\centering
\small
\setlength{\tabcolsep}{5.8pt}
\renewcommand{\arraystretch}{1.15}
\caption{CRMAgentBench Task Success Rate (TSR, \%), overall and per domain. Each
task is a full multi-turn conversation with an LLM user simulator over stateful
tools; TSR is a strict all-or-nothing pass. Overall is pooled across all five
domains (97 tasks total). Best in each column in bold; our model highlighted.
Frontier and larger-open baselines are shown for context.}
\label{tab:crmagentbench}
\begin{tabular*}{\textwidth}{@{\extracolsep{\fill}}lcccccc@{}}
\toprule
\textbf{Model}
& \shortstack{\textbf{Overall}\\\textbf{TSR}}
& \shortstack{\textbf{Store}\\\textbf{Manager}}
& \shortstack{\textbf{Customer}\\\textbf{Service}}
& \shortstack{\textbf{Field}\\\textbf{Service}}
& \textbf{Marketing}
& \textbf{SDR} \\
\midrule
GPT-4.1                    & 82 & 67 & 77 & 83 & \textbf{90} & 95 \\
\ours
Salesforce Koa (FP8)       & 87 & \textbf{86} & 77 & \textbf{91} & \textbf{90} & 84 \\
Nemotron-3-Super-120B (BF16) & 79 & 76 & 77 & 74 & 86 & 84 \\
GPT-5.5                    & \textbf{91} & 81 & \textbf{92} & \textbf{91} & \textbf{90} & \textbf{100} \\
Gemini 3.7 Flash           & \textbf{91} & \textbf{86} & \textbf{92} & \textbf{91} & \textbf{90} & 95 \\
Nemotron-Ultra-550B        & 87 & 76 & \textbf{92} & \textbf{91} & \textbf{90} & 84 \\
\bottomrule
\end{tabular*}
\end{table}

\paragraph{Training.}
Rollouts are generated against NeMo Gym resource servers spanning
\code{salesforce support}, \code{healthcare administration}, \code{real estate},
and a helper-free \code{workplace assistant} sandbox
(Section~\ref{sec:data-regimes}). Training runs on a Slurm cluster of
5$\times$NVIDIA~B200 nodes: four actor nodes run colocated rollout generation and
policy training, and one dedicated helper node serves the Nano-30B environment
model; the policy is sharded TP=8, EP=8, PP=1 with sequence parallelism. Full
infrastructure details are in Appendix~\ref{appx:infra}.

\paragraph{Baselines.}
We compare Koa with two proprietary frontier models (GPT-4.1 and GPT-5.5)
and with its open-weight base, Nemotron-3-Super-120B, focusing on models evaluated on our production benchmark; CRMAgentBench adds
Gemini and Nemotron-Ultra-550B for context. GPT-4.1 and the untuned Nemotron base
are the two references we care most about: GPT-4.1 is a widely deployed
proprietary baseline, and the untuned base isolates the effect of our
post-training.

\paragraph{Serving and precision.}
We serve Nemotron and our fine-tuned checkpoints via vLLM with tensor parallelism,
supporting up to 256K context. These checkpoints are evaluated in BF16, except the
production Koa model, which is served in FP8 for inference efficiency; the
corresponding BF16 checkpoint is reported as \textbf{Koa (BF16)} to isolate the
effect of quantization.

\subsection{Enterprise benchmark results}
\label{sec:internal-benchmarks}

We evaluate on three enterprise benchmarks, each targeting CRM agent behavior from
a different angle: \textbf{CRMAgentBench}, a multi-domain, multi-turn agent
benchmark scored by end-to-end task success; a \textbf{production tool-calling
benchmark}---an internal set with human-labeled and synthetic splits---which gives
our most decisive, controlled result; and \textbf{CRM Bench}, a multi-dimensional assessment of single-turn CRM
assistant behavior (reported with the public benchmarks in
Section~\ref{sec:public-benchmarks}, Table~\ref{tab:crmbench}).

\subsubsection{CRMAgentBench}
\label{sec:crmagentbench}

CRMAgentBench is a multi-node ReAct agent framework with an integrated benchmark
suite. It ships five production-style CRM agents---Customer Service, Store Manager,
Field Service, Sales Development Rep (SDR), and Marketing---spanning 97 tasks. Each
task is run as a full multi-turn conversation between two language models: an LLM
user simulator that plays the customer from a scripted persona, and the agent under
test. Tools are stateful, reading and writing a shared mock database, so the
benchmark verifies the agent's side effects rather than only its final message. The
headline metric is \textbf{Task Success Rate} (TSR), a strict, all-or-nothing pass
per task: a task counts as solved only when all of its asserted checks hold
(actions taken, resulting state, tool arguments, response content and order, and
avoidance of forbidden actions). We report TSR overall (pooled across all 97 tasks)
and per domain.

As shown in Table~\ref{tab:crmagentbench}, Koa reaches \textbf{87\%} overall TSR,
clearly above GPT-4.1 (82\%) and its untuned Nemotron base
(79\%), and matching Nemotron-Ultra-550B despite being far smaller. Koa trails the strongest
generalists (GPT-5.5 and Gemini 3.7 Flash, both 91\%) by a modest margin. The
gains over the base are largest in Field Service ($74\to91$) and Store Manager
($76\to86$).

\subsubsection{Production tool-calling benchmark}
\label{sec:prod-benchmark}

We benchmark the production model against frontier and open baselines on two
in-house evaluation sets: an internal human-labeled set and an internal synthetic
set judged by multiple LLMs. The production Koa model is our single-environment RL checkpoint
(step 100) after FP8 post-training quantization. Metrics are tool-name accuracy
(\textit{Name}), LLM-judged argument accuracy (\textit{Args}), full tool-call
success with both name and arguments correct (\textit{Succ}), and the
non-hallucination rate (\textit{NoHall}, predicted function within the allowed
set); higher is better throughout. We present two views of the same benchmark:
Table~\ref{tab:prod-compare} places Koa against frontier and open baselines, and
Table~\ref{tab:prod-ablation} isolates the effect of training data and precision
with architecture and recipe held fixed.

On the human-labeled set in Table~\ref{tab:prod-compare}, Koa outperforms GPT-4.1
on all four metrics. It achieves the highest non-hallucination rate of any model
(0.989), higher tool-name accuracy than all evaluated open-weight baselines (0.966),
and argument and success scores (0.851/0.848) ahead of
GPT-5.5 and both Gemini models. On the synthetic set the frontier models
lead on success (GPT-5.5 0.902, GPT-4.1 0.897) while Koa scores 0.847. The
training-data and precision ablation in Table~\ref{tab:prod-ablation} holds
architecture and recipe fixed to isolate in-domain RL and FP8 quantization; we
analyze it in Section~\ref{sec:analysis}.

\begin{table}[t]
\caption{Production tool-calling benchmark: Koa versus frontier and open baselines.
\textit{Name}: tool-name accuracy; \textit{Args}: LLM-judged argument accuracy;
\textit{Succ}: full tool-call success (name and args both correct);
\textit{NoHall}: non-hallucination rate; higher is better. Best in each column in
bold; our deployed model highlighted. Dashes mark configurations not evaluated on
that set. The training-data and precision ablation is in
Table~\ref{tab:prod-ablation}.}
\label{tab:prod-compare}
\centering
\small
\setlength{\tabcolsep}{6pt}
\renewcommand{\arraystretch}{1.15}
\begin{adjustbox}{max width=\textwidth}
\begin{tabular}{@{}l cccc c cccc@{}}
\toprule
& \multicolumn{4}{c}{\textbf{Human-labeled (internal)}}
& & \multicolumn{4}{c}{\textbf{Synthetic (multi-LLM-judged)}} \\
\cmidrule(lr){2-5}
\cmidrule(lr){7-10}
\textbf{Model}
& \textbf{Name} & \textbf{Args} & \textbf{Succ} & \textbf{NoHall}
&
& \textbf{Name} & \textbf{Args} & \textbf{Succ} & \textbf{NoHall} \\
\midrule
GPT-4.1
    & 0.963 & 0.782 & 0.782 & 0.974
    && 0.937 & \textbf{0.954} & 0.897 & 0.995 \\
\ours
\textbf{Salesforce Koa (FP8)}
    & 0.966 & \textbf{0.851} & \textbf{0.848} & \textbf{0.989}
    && 0.902 & 0.931 & 0.847 & 0.983 \\
Nemotron-3-Super-120B (BF16)
    & 0.940 & 0.819 & 0.819 & 0.983
    && 0.891 & 0.927 & 0.846 & 0.972 \\
GPT-5.5
    & \textbf{0.971} & 0.788 & 0.788 & 0.986
    && \textbf{0.946} & \textbf{0.954} & \textbf{0.902} & \textbf{0.998} \\
GPT-5.4-mini
    & 0.946 & 0.817 & 0.817 & 0.971
    && 0.922 & 0.933 & 0.866 & 0.991 \\
Gemini 3.1 Pro
    & \textbf{0.971} & 0.842 & 0.842 & 0.983
    && 0.937 & 0.901 & 0.866 & 0.976 \\
Gemini 3.5 Flash
    & 0.954 & 0.822 & 0.822 & 0.980
    && 0.839 & 0.846 & 0.790 & 0.872 \\
gpt-oss-120b
    & 0.928 & 0.825 & 0.819 & 0.943
    && -- & -- & -- & -- \\
Nemotron-Ultra-550B
    & 0.874 & 0.791 & 0.785 & 0.885
    && 0.943 & 0.932 & 0.888 & 0.986 \\
\bottomrule
\end{tabular}
\end{adjustbox}
\end{table}

\begin{table}[t]
\caption{Production tool-calling benchmark: controlled comparison of training data
and precision, with architecture and RL recipe held fixed. Rows progress from the
untuned base through public-data RL and in-domain RL to the deployed FP8 model;
metrics are as in Table~\ref{tab:prod-compare}. Best in each column in bold; the
deployed model highlighted.}
\label{tab:prod-ablation}
\centering
\small
\setlength{\tabcolsep}{6pt}
\renewcommand{\arraystretch}{1.15}
\begin{adjustbox}{max width=\textwidth}
\begin{tabular}{@{}l cccc c cccc@{}}
\toprule
& \multicolumn{4}{c}{\textbf{Human-labeled (internal)}}
& & \multicolumn{4}{c}{\textbf{Synthetic (multi-LLM-judged)}} \\
\cmidrule(lr){2-5}
\cmidrule(lr){7-10}
\textbf{Configuration}
& \textbf{Name} & \textbf{Args} & \textbf{Succ} & \textbf{NoHall}
&
& \textbf{Name} & \textbf{Args} & \textbf{Succ} & \textbf{NoHall} \\
\midrule
Nemotron-3-Super-120B (BF16)
    & 0.940 & 0.819 & 0.819 & 0.983
    && 0.891 & 0.927 & 0.846 & 0.972 \\
\quad + RL, public data
    & 0.954 & 0.825 & 0.825 & 0.983
    && 0.888 & \textbf{0.932} & 0.845 & 0.971 \\
Salesforce Koa (BF16)
    & 0.948 & 0.842 & 0.842 & 0.986
    && 0.884 & \textbf{0.932} & 0.845 & 0.971 \\
\ours
\textbf{Salesforce Koa (FP8)}
    & \textbf{0.966} & \textbf{0.851} & \textbf{0.848} & \textbf{0.989}
    && \textbf{0.902} & 0.931 & \textbf{0.847} & \textbf{0.983} \\
\bottomrule
\end{tabular}
\end{adjustbox}
\end{table}

\subsection{Public benchmark results}
\label{sec:public-benchmarks}

We next check that specializing on the enterprise distribution did not erode
general tool-use capability, using two public benchmarks and one internal
single-turn benchmark. \textbf{Tau2Bench}~\citep{barres2025tau2} measures
end-to-end multi-turn customer service across airline, retail, and telecom
(GPT-4.1 user simulator, four trials, $\mathrm{pass}^{\wedge}1$ averaged);
\textbf{BFCL}~\citep{patil2025bfcl} measures agentic tool use across multi-step
calling, web search, memory, and stateful tools; and \textbf{CRM Bench} is an
internal single-turn Salesforce/Agentforce benchmark scored along topic,
function-call, and free-text accuracy. The message is deliberately modest: here
Koa should \emph{preserve} general capability, not win.

\paragraph{Tau2Bench.}
On the task-weighted average in Table~\ref{tab:tau2}, Koa reaches 69.33, above
GPT-4.1 (54.48) and its Nemotron base (68.64), with its strongest domain in
retail (83.3). It trails GPT-5.5 on the overall task-weighted average.

\begin{table}[t]
\centering
\small
\setlength{\tabcolsep}{6pt}
\renewcommand{\arraystretch}{1.15}
\caption{Tau2Bench (multi-turn customer service). The average is task-weighted
(Airline: 50, Retail: 114, Telecom: 114). Best in each column in bold; our model highlighted.}
\label{tab:tau2}
\begin{tabular*}{\textwidth}{@{\extracolsep{\fill}}lcccc@{}}
\toprule
\textbf{Model} & \textbf{Airline} & \textbf{Retail} & \textbf{Telecom}
& \shortstack{\textbf{Weighted}\\\textbf{Avg.}} \\
\midrule
GPT-4.1                & 56.0 & 74.1 & 34.2 & 54.48 \\
\ours
Salesforce Koa (FP8)   & 60.5 & \textbf{83.3} & 59.2 & 69.33 \\
Nemotron-3-Super-120B (BF16) & 61.5 & 79.9 & 60.5 & 68.64 \\
GPT-5.5                & \textbf{62.5} & 81.6 & \textbf{95.8} & \textbf{83.99} \\
\bottomrule
\end{tabular*}
\end{table}

\paragraph{BFCL.}
Overall and by major category in Table~\ref{tab:bfcl}, Koa scores 63.87\% overall,
above GPT-4.1 (53.96\%) and essentially at parity with its Nemotron base
(64.73\%), while below GPT-5.5 (67.63\%). The near-flat aggregate reflects small,
offsetting category movements: Koa
posts the best Live tool-use score of any model, gains on Web Search, and is
essentially unchanged on Multi-Turn, while softening modestly on static Non-Live AST,
Memory, and irrelevance detection. No category collapses, so specialization preserved
general tool use.

\begin{table}[t]
\caption{BFCL overall accuracy and per-category accuracy (\%). Categories are the
official BFCL groupings, with detection split into relevance (Rel.) and
irrelevance (Irrel.). Best in each column in bold; our model highlighted.
GPT-4.1 figures are the public BFCL~V4
leaderboard scores for GPT-4.1-2025-04-14 (FC); the other rows are our own runs on
the same benchmark categories.}
\label{tab:bfcl}
\centering
\footnotesize
\setlength{\tabcolsep}{5pt}
\renewcommand{\arraystretch}{1.12}
\begin{adjustbox}{max width=\textwidth}
\begin{tabular}{lcccccccc}
\toprule
\textbf{Model} & \textbf{Overall} & \shortstack{\textbf{Non-Live}\\\textbf{AST}}
& \textbf{Live} & \shortstack{\textbf{Multi-}\\\textbf{Turn}}
& \shortstack{\textbf{Web}\\\textbf{Search}} & \textbf{Memory}
& \shortstack{\textbf{Rel.}\\\textbf{Det.}} & \shortstack{\textbf{Irrel.}\\\textbf{Det.}} \\
\midrule
GPT-4.1                & 53.96 & 82.79 & 69.95 & 38.88 & 68.00 & 23.87 & \textbf{87.50} & \textbf{86.52} \\
\ours
Salesforce Koa (FP8)   & 63.87 & 82.90 & \textbf{80.98} & 53.50 & 74.50 & 50.75 & 81.25 & 63.79 \\
Nemotron-3-Super-120B (BF16) & 64.73 & \textbf{86.17} & 80.38 & \textbf{54.12} & 71.50 & 52.90 & 68.75 & 71.67 \\
GPT-5.5                & \textbf{67.63} & 81.65 & 67.73 & 50.50 & \textbf{80.50} & \textbf{64.52} & 62.50 & 85.41 \\
\bottomrule
\end{tabular}
\end{adjustbox}
\end{table}

\paragraph{CRM Bench.}
On this internal single-turn benchmark in Table~\ref{tab:crmbench}, Koa scores 0.87
overall, above GPT-4.1 (0.81) and the base (0.84); its
function-call accuracy (0.85) ties GPT-4.1 for best and improves markedly over the
base (0.71). Taken together, the three benchmarks show that after enterprise
specialization Koa \emph{preserves} general tool-use capability---at parity with
its base and far ahead of GPT-4.1---rather than trading it away.

\begin{table}[t]
\centering
\small
\setlength{\tabcolsep}{6pt}
\renewcommand{\arraystretch}{1.15}
\caption{CRM Bench (internal single-turn Salesforce/Agentforce). Scored on topic,
function-call, and free-text accuracy; the average is dimension-weighted. Best in each column in bold; our
model highlighted.}
\label{tab:crmbench}
\begin{tabular*}{\textwidth}{@{\extracolsep{\fill}}lcccc@{}}
\toprule
\textbf{Model} & \textbf{Topic} & \textbf{Func.} & \textbf{Text}
& \shortstack{\textbf{Weighted}\\\textbf{Avg.}} \\
\midrule
GPT-4.1                & 0.98 & \textbf{0.85} & 0.60 & 0.81 \\
\ours
Salesforce Koa (FP8)   & 0.97 & \textbf{0.85} & 0.78 & 0.87 \\
Nemotron-3-Super-120B (BF16) & 0.97 & 0.71 & 0.85 & 0.84 \\
GPT-5.5                & \textbf{0.99} & 0.82 & \textbf{0.89} & \textbf{0.90} \\
\bottomrule
\end{tabular*}
\end{table}

\section{Analysis}
\label{sec:analysis}

We now examine the effect of in-domain adaptation (Section~\ref{sec:indomain}), how SFT and RL differ as
adaptation strategies (Section~\ref{sec:sftrl}), and how general tool use shifts
across our checkpoints and how FP8 deployment affects aggregate and per-category performance
(Section~\ref{sec:bfcl-tradeoff}).

\subsection{Effect of in-domain adaptation}
\label{sec:indomain}

Table~\ref{tab:prod-ablation} provides a controlled comparison between the ``Nemotron-3-Super-120B + RL, public
data'' checkpoint and Koa, which continues that checkpoint with a further RL stage
on in-domain data. With the base and RL recipe held fixed, the added in-domain
stage is the only difference. On the human-labeled enterprise benchmark, this in-domain
stage improves success and argument accuracy from $0.825/0.825$ to $0.842/0.842$,
and Salesforce Koa reaches $0.848/0.851$.

\subsection{SFT versus RL from an already post-trained base}
\label{sec:sftrl}

A natural alternative to RL is supervised fine-tuning. Starting from the same
post-trained Nemotron base, we compare SFT-only and RL-only checkpoints, with the
Tau2Bench, BFCL, and CRM Bench breakdowns in Appendix~\ref{appx:sft-rl}
(Tables~\ref{tab:tau2-ablation}--\ref{tab:crm-ablation}).
On Tau2Bench both improve modestly over the base (SFT 70.04, RL 69.41, base
68.64), with SFT strongest by a small margin. The clear distinction is on BFCL
multi-turn tool use: SFT slightly reduces accuracy ($54.12\!\to\!53.25$) whereas
RL improves it substantially to $59.50$, and SFT's gains concentrate in memory and
relevance detection that do not transfer to multi-turn performance. On the
predominantly single-turn CRM Bench, SFT is effective ($0.84\!\to\!0.88$), edging
RL ($0.86$).

Because enterprise workflows are multi-turn, we choose RL over SFT as the
adaptation \emph{method}. The deployed Koa applies this RL recipe in two stages,
public/synthetic followed by in-domain Salesforce (Section~\ref{sec:indomain}), and
is served in FP8; the RL-only checkpoint here isolates the method choice against
SFT. We treat this comparison as scoped to a base that is itself already
RL-post-trained, since from a pre-RL checkpoint the conclusion could differ.

\subsection{General tool use across our checkpoints and FP8 deployment}
\label{sec:bfcl-tradeoff}

\begin{table}[t]
\caption{BFCL across Koa's two-stage RL pipeline: the Nemotron-3-Super-120B base, a
first RL stage on public/synthetic data, a second RL stage on enterprise data (Koa),
and FP8 deployment. All checkpoints before FP8 deployment are evaluated in BF16.
Columns are BFCL Overall accuracy and per-category scores. The
full comparison including the SFT-only checkpoint is in
Appendix~\ref{appx:sft-rl} (Table~\ref{tab:bfcl-ablation}).}
\label{tab:bfcl-transition}
\centering
\footnotesize
\setlength{\tabcolsep}{5pt}
\renewcommand{\arraystretch}{1.12}
\begin{adjustbox}{max width=\textwidth}
\begin{tabular}{lcccccccc}
\toprule
\textbf{Setup} & \textbf{Acc.} & \shortstack{\textbf{Non-Live}\\\textbf{AST}} & \textbf{Live}
& \shortstack{\textbf{Multi-}\\\textbf{Turn}} & \shortstack{\textbf{Web}\\\textbf{Search}} & \textbf{Memory}
& \shortstack{\textbf{Rel.}\\\textbf{Det.}} & \shortstack{\textbf{Irrel.}\\\textbf{Det.}} \\
\midrule
Nemotron-3-Super-120B (BF16) & 64.73 & \textbf{86.17} & 80.38 & 54.12 & 71.50 & 52.90 & 68.75 & \textbf{71.67} \\
\quad + RL, public/synthetic data & \textbf{66.63} & 84.85 & 80.16 & \textbf{59.50} & 73.50 & 52.04 & 68.75 & 71.66 \\
Salesforce Koa (BF16) & 64.31 & 84.00 & 80.68 & 57.50 & 67.00 & \textbf{54.62} & 75.00 & 62.70 \\
\ours
Salesforce Koa (FP8) & 63.87 & 82.90 & \textbf{80.98} & 53.50 & \textbf{74.50} & 50.75 & \textbf{81.25} & 63.79 \\
\bottomrule
\end{tabular}
\end{adjustbox}
\end{table}

The BFCL breakdown in Table~\ref{tab:bfcl-transition} shows how general tool use
shifts across Koa's two-stage pipeline, and it clarifies where the headline
aggregate comes from. The first stage, RL on the public/synthetic regime, scores
highest on BFCL, above the Nemotron base ($66.63$ vs.\ $64.73$), with its largest
gains on multi-turn tool use ($54.12\!\to\!59.50$) and web search
($71.50\!\to\!73.50$); the spec-driven recipe generalizes well on public agentic
tasks. 
The second stage, RL on in-domain data, specializes the model to the production distribution and brings BFCL performance back near the base level ($66.63\!\to\!64.31$), with lower scores in web search ($67.00$).

FP8 deployment yields comparable aggregate benchmark performance. Comparing
the enterprise BF16 checkpoint with the deployed FP8 model, the Tau2Bench weighted
average changes only from $69.70$ to $69.33$, CRM Bench rises from $0.86$ to $0.87$,
and on the production benchmark FP8 with in-domain calibration is at or above BF16
on every human-labeled metric (Table~\ref{tab:prod-ablation}). On BFCL the aggregate
moves only $0.44$ points ($64.31\!\to\!63.87$), with lower Multi-Turn and Memory
scores offset by gains in Web Search and relevance detection. FP8 therefore
retains similar aggregate performance for 120B deployment, with category-level trade-offs.

Read alongside the production benchmark, this is a favorable trade: the second
in-domain stage costs $2.3$ BFCL points ($66.63\!\to\!64.31$) yet is the same stage
that lifts the deployed task enterprise use depends on (human-labeled success
$0.825\!\to\!0.842$, Table~\ref{tab:prod-ablation}; $0.848$ in FP8). BFCL peaks on
the public regime, whereas the deployed model optimizes the in-domain distribution
that drives enterprise value, holding parity with its base on public tool use while
leading where deployment matters.

\section{Conclusion}
\label{sec:conclusion}
We presented Salesforce Koa, an enterprise language model built by post-training
the open-weight Nemotron-120B foundation model with GRPO and deployed in FP8. On
CRMAgentBench and the human-labeled production tool-calling set, Koa outperforms
both its untuned open-weight base and GPT-4.1 and is competitive with the strongest
frontier models, while preserving general tool-use capability on public benchmarks.
Its central contribution is a specification-driven RL pipeline in which the same
declarative Agent Script specifications that configure an agent also supply the
workflow structure and grounded rewards for its rollout tasks; instantiated on
enterprise CRM specifications, this yields the in-domain distribution on which Koa
is specialized. 
Our controlled comparison shows that, with architecture and RL recipe held fixed, an additional in-domain RL stage improves argument accuracy and full tool-call success on the human-labeled enterprise benchmark.

Two limitations temper these results: our base is itself already RL-post-trained,
so the preference for RL over SFT may not hold from a pre-RL checkpoint, and some
enterprise evaluations use modest task counts. Future work includes co-designing
the SFT and RL stages, evaluating from a pre-RL foundation, and extending
specification-driven environments to broader domains and more complex agentic
workflows.

\section{Acknowledgements}
This model was trained in partnership with NVIDIA.

\bibliographystyle{plainnat}
\bibliography{\koaroot custom}

% ---- Appendices (one file per section) ------------------------------------
\clearpage
\appendix
\section{Environments and rollout}

\subsection{Environment interaction and verification}
\label{appx:env-summary}

Table~\ref{tab:env-summary} summarizes how the training environments
(Section~\ref{sec:environment}) differ in interaction topology and verification.

\begin{table}[H]
\centering
\caption{Environment interaction and verification mechanisms.}
\label{tab:env-summary}
\small
\renewcommand{\arraystretch}{1.12}
\begin{tabularx}{\textwidth}{@{}>{\raggedright\arraybackslash}p{2.6cm} Y Y@{}}
\toprule
\textbf{Environment type} & \textbf{Interaction topology} & \textbf{Verification} \\
\midrule
Flat customer support &
Single agent; all advertised tools directly callable; persona-conditioned
simulated customer. &
Per-sub-question LLM coverage with a hard duplicate-call gate; the judge prompt
enforces successful-tool grounding for data-bound questions. \\
Routed customer service &
Router plus specialized subagents reached through \code{go\_to\_*}; out-of-scope
calls return a wrong-subagent error; a focus system message is inserted on each
switch. &
Same coverage-based reward and duplicate-call gate, with domain-specific scope
and grounding rules in the judge prompt. \\
Deterministic tool sandbox &
Single agent over stateful tools; no simulated customer and no helper. &
Binary state-equivalence: reward is one only if the state produced by the
predicted actions matches the reference actions' state. \\
\bottomrule
\end{tabularx}
\end{table}

\subsection{Training rollout}
\label{appx:rollout}
\begin{figure}[H]
\centering
\begin{tikzpicture}[font=\small, yscale=0.9,
  msg/.style={-{Latex[length=2mm]}, thick, draw=black!70},
  ret/.style={-{Latex[length=2mm]}, semithick, dashed, draw=black!55},
  life/.style={draw=black!35, dashed},
  actor/.style={rectangle, rounded corners=2pt, draw=black!55, fill=black!3,
                align=center, minimum height=8mm, text width=25mm}]
  \node[actor] (P) at (0,0)    {Policy\\(trainable)};
  \node[actor] (A) at (3.7,0)  {Env agent\\+ router};
  \node[actor] (R) at (7.4,0)  {Resource\\server};
  \node[actor] (H) at (11.1,0) {Frozen helper\\(sim / emulator / judge)};
  \foreach \n in {P,A,R,H}{\draw[life] (\n.south) -- ($(\n.south)+(0,-9.6)$);}
  \draw[msg] (0,-1.1) -- node[above,font=\scriptsize]{response / tool call / \code{go\_to\_*}} (3.7,-1.1);
  \draw[msg] (3.7,-1.9) -- node[above,font=\scriptsize]{dispatch (scope + schema check)} (7.4,-1.9);
  \draw[msg] (7.4,-2.7) -- node[above,font=\scriptsize]{emulate tool result} (11.1,-2.7);
  \draw[ret] (11.1,-3.4) -- node[below,font=\scriptsize]{synthesized output} (7.4,-3.4);
  \draw[msg] (7.4,-4.2) -- node[above,font=\scriptsize]{tool result / observation} (0,-4.2);
  \draw[msg] (3.7,-5.2) -- node[above,font=\scriptsize]{request next customer turn} (11.1,-5.2);
  \draw[ret] (11.1,-5.9) -- node[below,font=\scriptsize]{simulated user message} (0,-5.9);
  \node[draw=black!30, fill=yellow!10, rounded corners=2pt, font=\scriptsize,
        align=center, inner sep=3pt] at (5.55,-6.7)
        {repeat until the session ends or the turn budget is reached};
  \draw[msg] (7.4,-7.7) -- node[above,font=\scriptsize]{coverage judge over sub-questions} (11.1,-7.7);
  \draw[ret] (11.1,-8.4) -- node[below,font=\scriptsize]{per-sub-question verdicts} (7.4,-8.4);
  \draw[msg] (7.4,-9.2) -- node[above,font=\scriptsize]{scalar reward + diagnostics} (3.7,-9.2);
\end{tikzpicture}
\caption{One training rollout. Solid arrows are the trained/rollout path; dashed
arrows are frozen-helper responses. The policy (left) is the only trained
component; the reward is produced by the coverage judge at the end.}
\label{fig:rollout}
\end{figure}
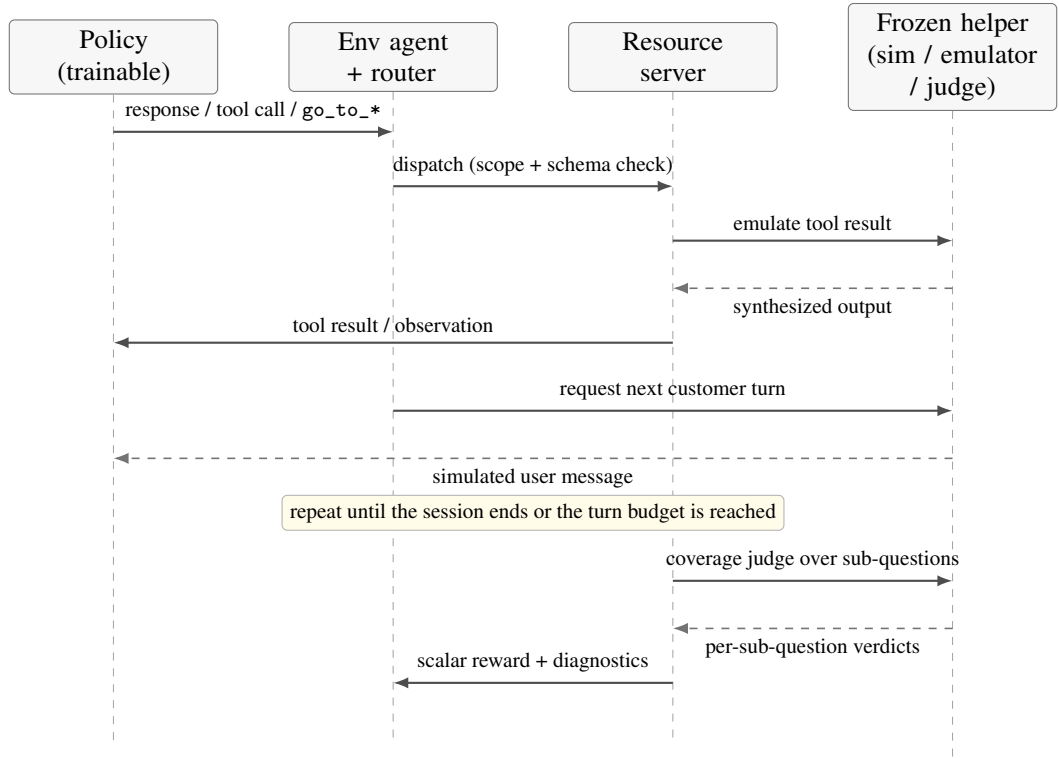

Figure~\ref{fig:rollout} traces one training rollout
(Section~\ref{sec:environment}) end to end, making explicit which components are
frozen versus trained, the order of interactions within a turn, and where the
scalar reward comes from.

\section{Training and evaluation details}

\subsection{GRPO optimization}
\label{appx:grpo}
We maximize $\mathcal{J}(\theta)=\mathbb{E}_{x\sim\mathcal{D}}\,
\mathbb{E}_{\tau\sim\pi_\theta(\cdot\mid x,\mathrm{env})}[R(\tau)]$, where $R(\tau)$
is the reward of Section~\ref{sec:environment}. Because the reward is available
only at trajectory end and we train no value network, advantages are estimated by
comparing trajectories sampled for the same prompt. Differential exposure across
environments is realized by repeating an environment's examples before shuffling,
$\mathcal{D}_{\mathrm{train}}=\operatorname{Shuffle}(\biguplus_{e\in\mathcal{E}}
r_e\,\mathcal{D}^{(e)}_{\mathrm{train}})$, with per-environment factors $r_e$; the
validation corpus is not upsampled.

\paragraph{Group and dynamic sampling.} For each prompt the policy samples a small
group of complete trajectories with colocated inference; groups with zero reward
variance carry no signal and are discarded, refilling until a full batch is
assembled.

\begin{samepage}
\paragraph{Leave-one-out normalized advantages.} For trajectory $i$ in a group of
$G$ with rewards $R_j$, with $b_i=\tfrac{1}{G-1}\sum_{j\ne i}R_j$ and
$s_i=\operatorname{std}\{R_j:j\ne i\}$, the advantage is
$A_i=\operatorname{clip}\!\big(\tfrac{R_i-b_i}{s_i+\epsilon},-c,c\big)$ for
$s_i>0$. When $s_i=0$, we skip normalization and clip the unnormalized advantage
$R_i-b_i$ to $[-c,c]$. $A_i$ is expanded over the generated tokens of trajectory $i$.
\par
\end{samepage}

\paragraph{On-policy update with sampling correction.} We perform exactly one
update per rollout batch with the policy ratio forced to one (no clipped-ratio
term, no reference-KL penalty); a token-level truncated importance-sampling weight
$w_{i,t}=\min(\tau,\exp[\log\pi_{\mathrm{train}}(y_{i,t})-\log\pi_{\mathrm{gen}}(y_{i,t})])$
corrects the train/generation mismatch. With sequence-level aggregation,
\begin{equation}
  \mathcal{L}_{\mathrm{GRPO}}(\theta)
  = -\frac{1}{B}\sum_{i=1}^{B}\frac{1}{|y_i|}\sum_{t=1}^{|y_i|}
    w_{i,t}\,A_i\,\log\pi_\theta(y_{i,t}\mid x_i,y_{i,<t}).
  \label{eq:grpo-loss}
\end{equation}
We average over valid assistant-generated tokens within each trajectory, then
over valid trajectories; user messages and tool responses serve only as
conditioning context.

\paragraph{Validity constraints.} Trajectories with malformed tool-call or thinking
syntax have the offending assistant-message token advantages set to a fixed
negative value; over-length or log-probability-inconsistent trajectories are
masked. A complementary single-step mode for \emph{constrained decision turns}
(Appendix~\ref{appx:constrained}) targets turns where the agent must communicate
with the user rather than call a tool.

\subsection{Constrained decision turns}
\label{appx:constrained}
A complementary training mode targets \emph{constrained decision turns}, where the
agent cannot resolve a request through tool execution alone and must instead
interact with the user (communicate a limitation, request missing information, or
manage expectations). We create the constraint dynamically by withholding a
needed tool at a designated turn while the simulated customer issues a request
that requires it. We form a single prompt-response training row at that turn.
The single-step reward follows the
same gate-then-score pattern as Eq.~\ref{eq:reward}: a procedural gate zeroes the
reward on any premature state-changing call, and a consequence score grades whether
the withheld tool is correctly emitted at a later recovery turn (generated only
for reward computation and excluded from the gradient)~\cite{chen2026criticalstate}.
These rows are trained with
the same GRPO optimizer (Eq.~\ref{eq:grpo-loss}); the environment builder produces
both turn-0 and mid-conversation constrained states.

\subsection{Distributed training and offline evaluation}
\label{appx:infra}
Policy training is distributed with tensor, expert, and sequence parallelism
(TP=8, EP=8, PP=1); rollout generation runs colocated on the actor workers,
alternating generation and optimization. The frozen environment helper is served
as a single separate inference engine to avoid a data-parallel coordination
failure mode observed when the helper was split across replicas. Single-env GRPO
uses 4 actor nodes (32 B200s, DP=4) plus 1 dedicated helper node (8 B200s, TP=8);
multi-env GRPO uses the same layout. During training, validation replays held-out
environment examples and reports aggregate and per-environment reward. Offline
evaluation is kept independent of the training-time helper: checkpoints are served
under greedy decoding, tool calls are parsed with a schema-aware parser, and
tool-name, argument, and full-call correctness are scored, including an independent
LLM judge for argument equivalence.

\begin{samepage}
\section{Additional post-training experiments}
\label{appx:sft-rl}

\subsection{Preliminary SFT setup}
\label{appx:sft}
\end{samepage}
We ran a preliminary SFT study on the already post-trained Nemotron base using
a masked next-token loss restricted to target (assistant and tool-call) tokens. The entire
corpus is synthetic, generated with a pipeline built on
APIGen~\cite{liu2024apigenautomatedpipelinegenerating} and
APIGen-MT~\cite{prabhakar2025apigenmt}, using only publicly available
function-calling resources (tool collections from BFCL~\cite{patil2025bfcl} and
Tau2Bench Airline~\cite{barres2025tau2}) as schemas and executable environments.
We perform full-parameter fine-tuning of
Nemotron-120B with NeMo Automodel~\cite{nemo-automodel} over 32 H200 GPUs (FSDP2,
expert parallelism, activation checkpointing) at sequence length $8192$ and global
batch size $32$, with a deliberately conservative recipe (peak learning rate
$5\times10^{-8}$, cosine decay, gradient clipping, early stopping) to avoid
catastrophic forgetting. The BFCL multi-turn score is essentially unchanged
($54.1\%\!\to\!53.2\%$; Table~\ref{tab:bfcl-ablation}), motivating RL applied
directly to the base model.

\subsection{Post-training and precision comparisons}
Tables~\ref{tab:tau2-ablation}--\ref{tab:crm-ablation} give the full per-benchmark
results behind Section~\ref{sec:sftrl} and Section~\ref{sec:bfcl-tradeoff}.
The SFT-only and public/synthetic RL checkpoints are trained separately from the
same Nemotron-3-Super-120B base. Koa (BF16) continues the public/synthetic RL
checkpoint with the enterprise-data RL stage, and Koa (FP8) is its quantized
deployment version. The BFCL table includes the SFT-only checkpoint alongside
the checkpoints in the main-body Table~\ref{tab:bfcl-transition}.

\begingroup
\setlength{\intextsep}{8pt plus 2pt minus 2pt}
\begin{table}[H]
\caption{Tau2Bench results across customer-service domains. Weighted average
accounts for the number of tasks per domain. All checkpoints are evaluated in
BF16 except Salesforce Koa (FP8).}
\label{tab:tau2-ablation}
\centering
\small
\setlength{\tabcolsep}{6pt}
\renewcommand{\arraystretch}{1.1}
\begin{tabular*}{\textwidth}{@{\extracolsep{\fill}}lcccc@{}}
\toprule
\textbf{Setup} & \textbf{Airline} & \textbf{Retail} & \textbf{Telecom} & \textbf{Weighted Avg.} \\
\midrule
Nemotron-3-Super-120B (BF16) & 61.50 & 79.90 & 60.50 & 68.64 \\
\quad + SFT, public/synthetic data     & \textbf{64.50} & 79.82 & 62.70 & \textbf{70.04} \\
\quad + RL, public/synthetic data      & 62.00 & 81.58 & 60.50 & 69.41 \\
Salesforce Koa (BF16)                 & \textbf{64.50} & 77.85 & \textbf{63.82} & 69.70 \\
\ours
Salesforce Koa (FP8)                   & 60.50 & \textbf{83.33} & 59.21 & 69.33 \\
\bottomrule
\end{tabular*}
\end{table}

\begin{table}[H]
\caption{Post-training and precision comparisons on BFCL, including the
SFT-only checkpoint alongside the RL pipeline in Table~\ref{tab:bfcl-transition}.
All checkpoints are evaluated in BF16 except Salesforce Koa (FP8).}
\label{tab:bfcl-ablation}
\centering
\footnotesize
\setlength{\tabcolsep}{5pt}
\renewcommand{\arraystretch}{1.12}
\begin{adjustbox}{max width=\textwidth}
\begin{tabular}{lcccccccc}
\toprule
\textbf{Setup} & \textbf{Acc.} & \shortstack{\textbf{Non-Live}\\\textbf{AST}} & \textbf{Live}
& \shortstack{\textbf{Multi-}\\\textbf{Turn}} & \shortstack{\textbf{Web}\\\textbf{Search}} & \textbf{Memory}
& \shortstack{\textbf{Rel.}\\\textbf{Det.}} & \shortstack{\textbf{Irrel.}\\\textbf{Det.}} \\
\midrule
Nemotron-3-Super-120B (BF16) & 64.73 & \textbf{86.17} & 80.38 & 54.12 & 71.50 & 52.90 & 68.75 & \textbf{71.67} \\
\quad + SFT, public/synthetic data     & 65.03 & 82.88 & 80.90 & 53.25 & 72.00 & \textbf{57.42} & \textbf{87.50} & 67.93 \\
\quad + RL, public/synthetic data      & \textbf{66.63} & 84.85 & 80.16 & \textbf{59.50} & 73.50 & 52.04 & 68.75 & 71.66 \\
Salesforce Koa (BF16)                 & 64.31 & 84.00 & 80.68 & 57.50 & 67.00 & 54.62 & 75.00 & 62.70 \\
\ours
Salesforce Koa (FP8)                   & 63.87 & 82.90 & \textbf{80.98} & 53.50 & \textbf{74.50} & 50.75 & 81.25 & 63.79 \\
\bottomrule
\end{tabular}
\end{adjustbox}
\end{table}

\begin{table}[H]
\caption{Internal CRM Bench results. All checkpoints are evaluated in BF16
except Salesforce Koa (FP8).}
\label{tab:crm-ablation}
\centering
\small
\setlength{\tabcolsep}{6pt}
\renewcommand{\arraystretch}{1.1}
\begin{tabular*}{\textwidth}{@{\extracolsep{\fill}}lcccc@{}}
\toprule
\textbf{Setup} & \textbf{Topic Acc.} & \textbf{Func. Acc.} & \textbf{Text Acc.} & \textbf{Average} \\
\midrule
Nemotron-3-Super-120B (BF16) & 0.970 & 0.710 & 0.850 & 0.84 \\
\quad + SFT, public/synthetic data     & \textbf{0.980} & 0.770 & \textbf{0.890} & \textbf{0.88} \\
\quad + RL, public/synthetic data      & 0.970 & 0.770 & 0.850 & 0.86 \\
Salesforce Koa (BF16)                 & \textbf{0.980} & 0.780 & 0.820 & 0.86 \\
\ours
Salesforce Koa (FP8)                   & 0.968 & \textbf{0.845} & 0.782 & 0.87 \\
\bottomrule
\end{tabular*}
\end{table}
\endgroup

\end{document}